\documentclass{article}

\usepackage{times}
\usepackage{graphicx} %
\usepackage{subfigure}
\usepackage{amsmath}
\usepackage{amssymb}
\usepackage{natbib}
\usepackage{easytable}

\usepackage{amsthm}

\newtheorem{theorem}{Theorem}

\graphicspath{{images/}}

\usepackage{algorithm}
\usepackage{algorithmic}

\DeclareMathOperator*{\argmin}{argmin}

\DeclareMathOperator*{\minimize}{minimize}
\DeclareMathOperator*{\subjectto}{subject\;to}

\newcommand{\dd}{\mathsf{d}}
\newcommand{\RR}{\mathbb{R}}

\usepackage{hyperref}

\usepackage[accepted]{icml2017}

\icmltitlerunning{OptNet: Differentiable Optimization as a Layer in Neural Networks}
\begin{document}

\twocolumn[
\icmltitle{OptNet: Differentiable Optimization as a Layer in Neural Networks}

\begin{icmlauthorlist}
\icmlauthor{Brandon Amos}{cmu}
\icmlauthor{J.~Zico Kolter}{cmu}
\end{icmlauthorlist}

\icmlaffiliation{cmu}{School of Computer Science,
  Carnegie Mellon University. Pittsburgh, PA, USA}

\icmlcorrespondingauthor{Brandon Amos}{bamos@cs.cmu.edu}
\icmlcorrespondingauthor{J.~Zico Kolter}{zkolter@cs.cmu.edu}

\icmlkeywords{deep learning, convex optimization, quadratic programming}

\vskip 0.3in
]

\printAffiliationsAndNotice{}  %
\setcounter{footnote}{1} %

\begin{abstract}
  This paper presents OptNet, a network architecture that integrates
  optimization problems (here, specifically in the form of quadratic programs)
  as individual layers in larger end-to-end trainable deep networks.
  These layers encode constraints and complex dependencies
  between the hidden states that traditional convolutional and
  fully-connected layers often cannot capture.
  We explore the foundations for such an architecture:
  we show how techniques from sensitivity analysis, bilevel
  optimization, and implicit differentiation can be used to
  exactly differentiate through these layers and with respect
  to layer parameters;
  we develop a highly efficient solver for these layers that exploits fast
  GPU-based batch solves within a primal-dual interior point method, and which
  provides backpropagation gradients with virtually no additional cost on top of
  the solve;
  and we highlight the application of these approaches in several problems.
  In one notable example, the method is
  learns to play mini-Sudoku (4x4) given just input and output games,
  with no a-priori information about the rules of the game;
  this highlights the ability of OptNet to learn hard
  constraints better than other neural architectures.
\end{abstract}

\section{Introduction}

In this paper, we consider how to treat exact, constrained optimization as
an individual layer within a deep learning architecture.
Unlike traditional feedforward networks, where the output of each
layer is a relatively simple (though non-linear) function of the previous layer,
our optimization framework allows for individual layers to capture much richer
behavior, expressing complex operations that in total can reduce the overall
depth of the network while preserving richness of representation.  Specifically,
we build a framework where the output of the $i+1$th layer in a network is the
\emph{solution} to a constrained optimization problem based upon previous
layers.  This framework naturally encompasses a wide variety of inference
problems expressed within a neural network, allowing for the potential of much
richer end-to-end training for complex tasks that require such inference
procedures.

Concretely, in this paper we specifically consider the task of
solving small quadratic programs as individual layers.
These optimization problems are well-suited to capturing interesting
behavior and can be efficiently solved with GPUs.
Specifically, we consider layers of the form
\begin{equation}
\begin{split}
z_{i+1} = \argmin_{z} \;\; & \frac{1} {2}z^T Q(z_i) z + q(z_i)^T z \\
\subjectto \;\; & A(z_i) z  = b(z_i) \\
& G(z_i) z \leq h(z_i)
\end{split}
\label{eq:qp}
\end{equation}
where $z$ is the optimization variable, $Q(z_i)$, $q(z_i)$, $A(z_i)$, $b(z_i)$,
$G(z_i)$, and $h(z_i)$ are parameters of the optimization problem.
As the notation suggests, these parameters can depend in any differentiable way
on the previous layer $z_i$, and which can eventually be optimized just like
any other weights in a neural network.  These layers can be learned by taking
the gradients of some loss function with respect to the parameters.
In this paper, we derive the gradients of \eqref{eq:qp} by taking
matrix differentials of the KKT conditions of the optimization
problem at its solution.

In order to the make the approach practical for larger
networks, we develop a custom solver which can simultaneously solve multiple
small QPs in batch form.  We do so by developing a custom primal-dual
interior point method tailored specifically to dense batch operations on a GPU.
In total, the solver can solve batches of quadratic programs over 100 times
faster than
existing highly tuned quadratic programming solvers such as Gurobi and CPLEX.
One crucial algorithmic insight in the solver is that by using a
specific factorization of the primal-dual interior point update, we can obtain a
backward pass over the optimization layer virtually ``for free''
(i.e., requiring no additional factorization once the optimization problem itself
has been solved).
Together, these innovations enable parameterized optimization problems
to be inserted within the architecture of existing deep networks.

\newpage
We begin by highlighting background and related work, and then present our
optimization layer.  Using matrix differentials we derive rules for
computing the backpropagation updates.  We then present our
solver for these quadratic programs, based upon a
state-of-the-art primal-dual interior point method, and highlight the
novel elements as they apply to our formulation, such as
the aforementioned fact
that we can compute backpropagation at very little additional cost.
We then provide experimental results that demonstrate the capabilities of the
architecture, highlighting potential tasks that these architectures can solve,
and illustrating improvements upon existing approaches.

\section{Background and related work}
Optimization plays a key role in modeling complex phenomena and providing
concrete decision making processes in sophisticated environments.  A full
treatment of optimization applications is beyond our scope
\cite{boyd2004convex} but these methods have bound applicability in
control frameworks \cite{morari1999model,sastry2011adaptive}; numerous statistical
and mathematical formalisms \cite{sra2012optimization}, and physical
simulation problems like rigid body dynamics \cite{lotstedt1984numerical}.
Generally speaking, our work is a step towards learning optimization problems
behind real-world processes from data that can be learned
end-to-end rather than requiring human specification
and intervention.

In the machine learning setting, a wide array of applications consider
optimization as a means to perform inference in learning.
Among many other applications, these architectures are well-studied for
generic classification and structured prediction tasks
\citep{goodfellow2013multi,stoyanov2011empirical,brakel2013training,lecun2006tutorial,belanger2016structured,belanger2017end,amos2017input};
in vision for tasks such as denoising
\citep{tappen2007learning,schmidt2014shrinkage};
and \citet{metz2016unrolled} uses unrolled optimization within a network to
stabilize the convergence of generative adversarial networks
\cite{goodfellow2014generative}.  Indeed, the general idea of solving
restricted classes of optimization problem using neural networks goes back
many decades \cite{kennedy1988neural, lillo1993solving}, but has seen a
number of advances in recent years.  These models are often trained by one of
the following four methods. 

\paragraph{Energy-based learning methods}
These methods can be used for tasks like (structured) prediction
where the training method shapes the energy function to be low around
the observed data manifold and high elsewhere
\cite{lecun2006tutorial}.
In recent years, there has been a strong push to further incorporate
structured prediction methods like conditional random fields as the
``last layer'' of a deep network architecture
\cite{peng2009conditional,zheng2015conditional,chen2015learning}
as well as in deeper energy-based architectures
\cite{belanger2016structured,belanger2017end,amos2017input}.
Learning in this context requires observed data, which isn't present in
some of the contexts we consider in this paper, and also
may suffer from instability issues when combined with
deep energy-based architectures as observed in
\citet{belanger2016structured,belanger2017end,amos2017input}.

\paragraph{Analytically}
If an analytic solution to the argmin can be found,
such as in an unconstrained quadratic minimization,
the gradients can often also be computed analytically.
This is done in \citet{tappen2007learning,schmidt2014shrinkage}.
We cannot use these methods for
the constrained optimization problems
we consider in this paper because
there are no known analytic solutions.

\paragraph{Unrolling}
The argmin operation over an unconstrained objective can be approximated
by a first-order gradient-based method and unrolled.
These architectures typically introduce an optimization
procedure such as gradient descent into the inference procedure.
This is done in
\citet{domke2012generic,amos2017input,belanger2017end,metz2016unrolled,goodfellow2013multi,stoyanov2011empirical,brakel2013training}.
The optimization procedure is unrolled automatically or manually
\cite{domke2012generic} to obtain derivatives during training that incorporate
the effects of these in-the-loop optimization procedures.
However, unrolling the computation of a method like gradient descent
typically requires a substantially larger network, and adds substantially
to the network's computational complexity.

In all of these existing cases, the optimization problem is unconstrained
and unrolling gradient descent is often easy to do.
When constraints are added to the optimization problem, iterative
algorithms often use a projection operator that may be difficult
to unroll through.
In this paper, we do \textbf{not} unroll an optimization procedure
but instead use argmin differentiation as described in the
next section.

\paragraph{Argmin differentiation}
\label{sec:rw:argmin-diff}
Most closely related to our own work, there have been several papers that
propose some form of differentiation through argmin operators.
These techniques also come up in bilevel optimization
\cite{gould2016differentiating,kunisch2013bilevel}
and sensitivity analysis
\cite{bertsekas1999nonlinear,fiacco1990sensitivity,bonnans2013perturbation}.
In the case of \citet{gould2016differentiating}, the authors describe general
techniques for differentiation through optimization problems, but only describe
the case of exact equality constraints rather than both equality and inequality
constraints (in the case inequality constraints, they add these via a barrier
function).
\citet{amos2017input} considers argmin differentiation within the context
of a specific optimization problem (the bundle method) but does not consider
a general setting. \citet{johnson2016composing} performs implicit differentiation on
(multi-)convex objectives with coordinate subspace constraints,
but don't consider inequality constraints and don't consider in
detail general linear equality constraints.
Their optimization problem is only in the final layer of a
variational inference network while we propose to insert optimization
problems anywhere in the network.
Therefore a special case of OptNet layers (with no inequality constraints)
has a natural interpretation in terms of Gaussian inference,
and so Gaussian graphical models (or CRF ideas more generally)
provide tools for making the computation more efficient and interpreting
or constraining its structure.
Similarly, the older work of \citet{mairal2012task} considered argmin
differentiation for a LASSO problem, deriving specific rules for this case, and
presenting an efficient algorithm based upon our ability to solve the LASSO
problem efficiently.

In this paper, we use implicit differentiation
\cite{dontchev2009implicit,griewank2008evaluating}
and techniques from matrix differential calculus \cite{magnus1988matrix}
to derive the gradients from the KKT matrix of the problem.
A notable difference from other work within ML that we are
aware of, is that we analytically differentiate through inequality as well as
just equality constraints by differentiating the complementarity conditions;
this differs from e.g., \citet{gould2016differentiating} where they instead
approximately convert the problem to an unconstrained one via a barrier method.
We have also developed methods to make this approach practical and reasonably
scalable within the context of deep architectures.

\section{OptNet: solving optimization within a neural network}
Although in the most general form, an OptNet layer can be any
optimization problem, in this paper we will study OptNet layers defined by a
quadratic program
\begin{equation}
\begin{split}
\minimize_{z} \;\; & \frac{1} {2}z^T Q z + q^T z \\
\subjectto \;\; & A z = b, \; G z \leq h
\end{split}
\label{eq:qp2}
\end{equation}
where $z \in \mathbb{R}^n$ is our optimization variable
$Q \in \mathbb {R}^{n \times n} \succeq 0$
(a positive semidefinite matrix),
$q \in \mathbb {R}^n$, $A\in \mathbb{R}^{m \times n}$,
$b \in \mathbb{R}^m$,
$G \in \mathbb{R}^ {p \times n}$ and
$h \in \mathbb{R}^{p}$ are problem data,
and leaving out the dependence on the
previous layer $z_i$ as we showed in \eqref{eq:qp}
for notational convenience.
As is well-known,
these problems can be solved in polynomial time using a variety of methods; if
one desires exact (to numerical precision) solutions to these problems, then
primal-dual interior point methods, as we will use in a later section, are the
current state of the art in solution methods.
In the neural network setting, the \emph{optimal solution} (or more generally,
a \emph{subset of the optimal solution}) of this optimization problems becomes
the output of our layer, denoted $z_{i+1}$, and any of the problem data
$Q, q, A, b, G, h$
can depend on the value of the previous layer $z_i$.
The forward pass in our OptNet architecture thus involves simply setting up
and finding the solution to this optimization problem.

Training deep architectures, however, requires that we not just have a forward
pass in our network but also a backward pass. This requires that we compute the
derivative of the solution to the QP with respect to its input parameters,
a general topic we topic we discussed previously.
To obtain these derivatives, we differentiate the KKT conditions
(sufficient and necessary conditions for optimality) of \eqref{eq:qp2} at a
solution to the problem using techniques
from matrix differential calculus \cite{magnus1988matrix}.
Our analysis here can be extended to
more general convex optimization problems.

The Lagrangian of \eqref{eq:qp2} is given by
\begin{equation}
L(z,\nu,\lambda)=\frac{1}{2}z^TQz+q^Tz+\nu^T(Az-b)+\lambda^T(Gz-h)
\end{equation}
where $\nu$ are the dual variables on the equality constraints
and $\lambda\geq 0$ are the dual variables on the inequality constraints.
The KKT conditions for stationarity, primal feasibility,
and complementary slackness are
\begin{equation}
\begin{split}
Qz^\star+q+A^T\nu^\star+G^T\lambda^\star &= 0 \\
Az^\star-b &= 0 \\
D(\lambda^\star)(Gz^\star-h) &= 0,
\end{split}
\end{equation}
where $D(\cdot)$ creates a diagonal matrix from a vector
and $z^\star$, $\nu^\star$ and $\lambda^\star$ are the optimal
primal and dual variables.
Taking the differentials of these conditions gives the equations
\begin{equation}
\begin{split}
\dd Qz^\star + Q \dd z + \dd q + \dd A^T \nu^\star + & \\
A^T \dd \nu + \dd G^T
\lambda^\star + G^T \dd \lambda & = 0 \\
\dd A z^\star + A \dd z - \dd b & = 0 \\
D(Gz^\star -h)\dd \lambda + D(\lambda^\star)(\dd G z^\star  + G \dd z  - \dd h)
& = 0
\end{split}
\end{equation}
or written more compactly in matrix form
\begin{equation}
  \begin{split}
\begin{bmatrix}
Q & G^T & A^T \\
D(\lambda^\star)G  & D(Gz^\star-h) & 0 \\
A & 0 & 0 \\
\end{bmatrix}
\begin{bmatrix}
\dd z \\
\dd \lambda \\
\dd \nu \\
\end{bmatrix} = \\
-
\begin{bmatrix}
\dd Qz^\star + \dd q + \dd G^T\lambda^\star + \dd A^T\nu^\star \\
D(\lambda^\star)\dd Gz^\star - D(\lambda^\star)\dd h \\
\dd Az^\star - \dd b \\
\end{bmatrix}.
  \end{split}
  \label{eq:kkt-diff}
\end{equation}
Using these equations, we can form the Jacobians of $z^\star$ (or
$\lambda^\star$ and $\nu^\star$, though we don't consider this case here), with
respect to any of the data parameters.  For example, if we wished to compute the
Jacobian $\frac{\partial z^\star}{\partial b} \in \mathbb{R}^{n \times m}$, we
would simply substitute $\dd b = I$ (and set all other differential terms in
the right hand side to zero), solve the equation, and the resulting value of
$\dd z$ would be the desired Jacobian.

In the backpropagation algorithm, however, we never want to explicitly form the
actual Jacobian matrices, but rather want to form the left matrix-vector product
with some previous backward pass vector $\frac{\partial \ell}{\partial z^\star}
\in \mathbb{R}^n$, i.e., $\frac{\partial \ell}{\partial z^\star} \frac {\partial
z^\star}{\partial b}$.   We can do this efficiently by noting the
solution for the $(\dd z, \dd \lambda, \dd \nu)$ involves multiplying the \emph
{inverse} of the left-hand-side matrix in \eqref{eq:kkt-diff} by some right hand
side.  Thus, if we multiply the backward pass vector by the transpose of the
differential matrix
\begin{equation}
\label{eq-d-def}
\begin{bmatrix}
d_z \\ d_\lambda \\ d_\nu
\end{bmatrix}
=
-
\begin{bmatrix}
Q & G^T D(\lambda^\star) & A^T \\
G  & D(Gz^\star-h) & 0 \\
A & 0 & 0 \\
\end{bmatrix}^{-1}
\begin{bmatrix}
\left(\frac{\partial \ell}{\partial z^\star}\right)^T \\ 0 \\ 0
\end{bmatrix}
\end{equation}
then the relevant gradients with respect to all the QP parameters can be given by
\begin{equation}
  \hspace{-4mm}
  \begin{aligned}
    \nabla_Q \ell &= \frac{1}{2}(d_z z^{\star T} + z^\star d_z^T) &
    \nabla_q \ell &= d_z \\
    \nabla_A \ell &= d_\nu z^{\star T} + \nu^\star d_z^T &
    \nabla_b \ell &= -d_\nu \\
    \nabla_G \ell &= D(\lambda^\star)d_\lambda z^{\star T} + \lambda^\star d_z^T &
    \nabla_h \ell &= -D(\lambda^\star) d_\lambda
  \end{aligned}
  \label{eq:grads}
\end{equation}
where as in standard backpropagation, all these terms are at most the size of
the parameter matrices.
Some of these parameters should depend on the previous layer
$z_i$ and the gradients with respect to the previous layer can
be obtained through the chain rule.
In the next section, we show that the solution to an
interior point method provides a factorization we can use to
compute these gradient efficiently.

\subsection{An efficient batched QP solver}
\label{sec:qp-solver}

Deep networks are typically trained in mini-batches to take advantage
of efficient data-parallel GPU operations.
Without mini-batching on the GPU, many modern deep learning
architectures become intractable for all practical purposes.
However, today's state-of-the-art QP solvers like Gurobi and CPLEX
do not have the capability of solving multiple optimization
problems on the GPU in parallel across the entire minibatch.
This makes larger OptNet layers become quickly intractable
compared to a fully-connected layer with the same number of parameters.

To overcome this performance bottleneck in our quadratic program,
we implemented a GPU-based primal-dual interior point
method (PDIPM) based on \citet{mattingley2012cvxgen}
that solves a batch of quadratic programs, and which provides the necessary
gradients needed to train these in an end-to-end fashion.
Our performance experiments in Section~\ref{sec:qp-timing} shows
that our solver is significantly faster than the standard
non-batch solvers Gurobi and CPLEX.

Following the method of \citet{mattingley2012cvxgen},
our solver introduces slack variables on the inequality constraints
and iteratively minimizes the residuals from the KKT conditions
over the primal variable $z\in\RR^n$, slack variable $s\in\RR^p$,
and dual variables
$\nu\in\RR^m$ associated with the equality constraints and
$\lambda\in\RR^p$ associated with the inequality constraints.
Each iteration computes the affine scaling directions by solving
\begin{equation}
  K
  \begin{bmatrix}
    \Delta z^{\rm aff} \\
    \Delta s^{\rm aff} \\
    \Delta \lambda^{\rm aff} \\
    \Delta \nu^{\rm aff} \\
  \end{bmatrix}
  =
  \begin{bmatrix}
    -(A^T\nu + G^T\lambda + Qz + q) \\
    -S\lambda \\
    -(Gz+s-h) \\
    -(Az-b) \\
  \end{bmatrix}
  \label{eq:cvxgen:affine}
\end{equation}
where
\begin{equation*}
  K =
  \begin{bmatrix}
    Q & 0 & G^T & A^T \\
    0 & D(\lambda) & D(s) & 0 \\
    G & I & 0 & 0 \\
    A & 0 & 0 & 0 \\
  \end{bmatrix},
\end{equation*}
then centering-plus-corrector directions by solving
\begin{equation}
  K
  \begin{bmatrix}
    \Delta z^{\rm cc} \\
    \Delta s^{\rm cc} \\
    \Delta \lambda^{\rm cc} \\
    \Delta \nu^{\rm cc} \\
  \end{bmatrix}
  =
  \begin{bmatrix}
    0 \\
    \sigma\mu 1 - D(\Delta s^{\rm aff}) \Delta \lambda^{\rm aff} \\
    0 \\
    0 \\
  \end{bmatrix},
  \label{eq:cvxgen:cc}
\end{equation}
where $\mu=s^T\lambda/p$ is the duality gap
and $\sigma$ is defined in \citet{mattingley2012cvxgen}.
Each variable $v$ is updated with
$\Delta v = \Delta v^{\rm  aff} + \Delta v^{\rm cc}$
using an appropriate step size.  We actually solve a symmetrized version of the
KKT conditions, obtained by scaling the second row block by $D(1/s)$.
We analytically decompose these systems into smaller
symmetric systems and pre-factorize portions of them
that don't change (i.e. that don't involve $D(\lambda/s)$
between iterations). We have implemented a batched version of this method with the
PyTorch library\footnote{\url{https://pytorch.org}} and have released
it as an open source library
at \url{https://github.com/locuslab/qpth}.
It uses a custom CUBLAS extension to compute batch
matrix factorizations and solves in parallel and provides the necessary
derivatives for end-to-end learning.

\subsubsection{Efficiently computing gradients}
\label{sec:qp-solver-grads}
The backward pass gradients can be computed ``for free''
after solving the original QP with
this primal-dual interior point method,
without an additional matrix factorization
or solve. Each iteration
cmputes an LU decomposition of the matrix $K_{\mathrm{sym}}$,\footnote{We
perform an LU decomposition of a subset of the matrix formed
by eliminating variables to create only a $p \times p$ matrix (the number of
inequality constraints) that needs to be factor during each iteration of the
primal-dual algorithm, and one $m \times m$ and one $n \times n$ matrix once at
the start of the primal-dual algorithm, though we omit the detail here.  We also
use an LU decomposition as this routine is provided in batch form by CUBLAS, but
could potentially use a (faster) Cholesky factorization if and when the
appropriate functionality is added to CUBLAS).}
which is a
symmetrized version of the matrix needed for computing the backpropagated
gradients.
We compute the $d_{z,\lambda,\nu}$ terms by solving
the linear system
\begin{equation}
  K_{\rm sym}
  \begin{bmatrix}
    d_z \\
    d_s \\
    \tilde{d}_\lambda \\
    d_\nu \\
  \end{bmatrix}
  =
  \begin{bmatrix}
    \left(-\frac{\partial \ell}{\partial z_{i+1}}\right)^T \\
    0 \\
    0 \\
    0 \\
  \end{bmatrix},
  \label{eq:grads-system}
\end{equation}
where $\tilde{d}_\lambda = D(\lambda^\star) d_\lambda$ for $d_\lambda$ as
defined in \eqref{eq-d-def}.  Thus, all the backward pass gradients can be computed
using the factored KKT matrix at the solution.  Crucially, because the
bottleneck of solving this linear system is computing the factorization of the
KKT matrix (cubic time as opposed to the quadratic time for solving via
backsubstitution once the factorization is computed), the additional time
requirements for computing all the necessary gradients in the backward pass is
virtually nonexistent compared with the time of computing the solution.  To the
best of our knowledge, this is the first time that this fact has been exploited
in the context of learning end-to-end systems.

\subsection{Properties and representational power}
\label{sec:rep-power}
In this section we briefly highlight some of the mathematical properties of
OptNet layers.  The proofs here are straightforward, and are mostly based upon
well-known results in convex analysis, so are deferred to the appendix.  The
first result simply highlights that (because the solution of
strictly convex QPs is continuous), that OptNet layers are subdifferentiable
everywhere, and differentiable at all but a measure-zero set of points.

\begin{theorem}
  \label{theorem:existence}
  Let $z^\star(\theta)$ be the output of an OptNet layer, where $\theta =
  \{Q,p,A,b,G,h\}$.  Assuming $Q \succ 0$ and that $A$ has full row rank, then
  $z^\star(\theta)$ is subdifferentiable 
  everywhere: $\partial z^\star(\theta) \neq \varnothing$, where $\partial
  z^\star(\theta)$ denotes the Clarke generalized subdifferential
  \cite{clarke1975generalized} (an extension of the subgradient to non-convex
  functions), and has a single unique element (the Jacobian) for all but a
  measure zero set of points $\theta$.
\end{theorem}

The next two results show the representational power of the OptNet layer,
specifically how an OptNet layer compares to the common linear layer followed by
a ReLU activation.  The first theorem shows that an OptNet layer can
approximate arbitrary elementwise piecewise-linear functions, and so among other
things can represent a ReLU layer.

\begin{theorem}
  \label{theorem:pwlinear}
Let $f: \mathbb{R}^n \rightarrow \mathbb{R}^n$ be an elementwise piecewise
linear function with $k$ linear regions.  Then the function can be represented
as an OptNet layer using $O(nk)$ parameters.  Additionally, the layer $z_{i+1} =
\max\{Wz_i + b, 0\}$ for $W \in \mathbb{R}^{n \times m}, b \in \mathbb{R}^n$ can
be represented by an OptNet layer with $O(mn)$ parameters. 
\end{theorem}

Finally, we show that the converse does not hold: that there are function
representable by an OptNet layer which cannot be represented exactly by a
two-layer ReLU layer, which take exponentially many units to approximate
(known to be a universal function approximator).  A simple example
of such a layer (and one which we use in the proof) is just the max over three
linear functions  $f(z) = \max \{a_1^T x, a_2^Tx, a_3^T x\}$.

\begin{theorem}
  \label{theorem:rep}
  Let $f(z) : \mathbb{R}^{n} \rightarrow \mathbb{R}$ be a scalar-valued function
  specified by an OptNet layer with $p$ parameters.  Conversely, let
  $f'(z) = \sum_{i=1}^m w_i \max\{a_i^Tz + b_i, 0\}$ be the output of a
  two-layer ReLU network.
  Then there exist functions that the ReLU network cannot represent
  exactly over all of $\mathbb{R}$, and which require $O(c^p)$ parameters to
  approximate over a finite region.
\end{theorem}

\subsection{Limitations of the method}
Although, as we will show shortly, the OptNet layer has several strong points,
we also want to highlight the potential drawbacks of this approach.
First, although, with an efficient batch solver, integrating an OptNet layer into
existing deep learning architectures is potentially practical, we do note that
solving optimization problems exactly as we do here has has cubic complexity in
the number of variables and/or constraints.  This contrasts with the quadratic
complexity of standard feedforward layers.  This means that we \emph{are}
ultimately limited to settings where the number of hidden variables in an OptNet
layer is not too large (less than 1000 dimensions seems to be the limits of what
we currently find to the be practical, and substantially less if one wants
real-time results for an architecture).

Secondly, there are many improvements to the OptNet layers that are still
possible.  Our QP solver, for instance, uses fully dense matrix operations,
which makes the solves very efficient for GPU solutions, and which also makes
sense for our general setting where the coefficients of the quadratic problem
can be learned.  However, for setting many real-world optimization problems
(and hence for architectures that wish to more closely mimic some real-world
optimization problem), there is often substantial structure (e.g., sparsity), in
the data matrices that can be exploited for efficiency.
There is of course no prohibition of incorporating sparse matrix methods into
the fast custom solver, but doing so would require substantial added complexity,
especially regarding efforts like finding minimum fill orderings for different
sparsity patterns of the KKT systems.
In our solver \verb!qpth!, we have started experimenting
with cuSOLVER's batched sparse QR factorizations and solves.

Lastly, we note that while the OptNet layers can be trained just as any neural
network layer, since they are a new creation and since they have manifolds in
the parameter space which have no effect on the resulting solution (e.g.,
scaling the rows of a constraint matrix and its right hand side does not change
the optimization problem), there is admittedly more tuning required to get these
to work.
This situation is common when developing new neural
network architectures and has also been reported in the
similar architecture of \citet{schmidt2014shrinkage}.
Our hope is that techniques for overcoming some of the challenges
in learning these layers will continue to be developed in future work.

\section{Experimental results}
In this section, we present several experimental results that highlight the
capabilities of the QP OptNet layer.
Specifically we look at
1) computational efficiency over exiting solvers;
2) the ability to improve upon existing convex problems such as those used
in signal denoising;
3) integrating the architecture into an generic deep learning architectures;
and 4) performance of our approach on a problem that is challenging for current approaches.
In particular, we want to emphasize the results of our system on learning
the game of (4x4) mini-Sudoku, a well-known logical puzzle;
our layer is able to directly learn the necessary constraints
using just gradient information and no a priori knowledge
of the rules of Sudoku.
The code and data for our experiments are open sourced in the
\verb!icml2017! branch of \url{https://github.com/locuslab/optnet}
and our batched QP solver is available as a library at
\url{https://github.com/locuslab/qpth}.

\subsection{Batch QP solver performance}
\label{sec:qp-timing}

All of the OptNet performance results in this section are run on
an unloaded Titan X GPU. Gurobi is run on an unloaded quad-core
Intel Core i7-5960X CPU @ 3.00GHz.

\begin{figure}[t]
  \centering
  \includegraphics[width=0.45\textwidth]{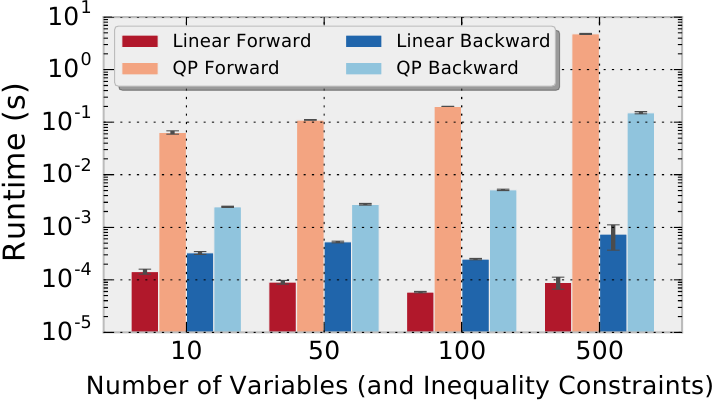}
  \caption{Runtime of a linear layer and a QP layer, batch of 128.}
  \label{fig:qp-perf-linear}
\end{figure}

Our OptNet layers are much more computationally expensive than
a linear or convolutional layer and a natural question is to ask
what the performance difference is.
We set up an experiment comparing a linear layer to a QP OptNet layer
with a mini-batch size of 128 on CUDA with
randomly generated input vectors sized 10, 50, 100, and 500.
Each layer maps this input to an output of the same dimension;
the linear layer does this with a batched matrix-vector multiplication
and the OptNet layer does this by taking the argmin of a random QP that
has the same number of inequality constraints as the dimensionality
of the problem.
Figure~\ref{fig:qp-perf-linear} shows the profiling results
(averaged over 10 trials) of the forward and backward passes.
The OptNet layer is significantly slower than the linear layer
as expected, yet still tractable in many practical contexts.

Our next experiment illustrates why standard baseline QP solvers
like CPLEX and Gurobi without batch support are too
computationally expensive for QP OptNet layers to be tractable.
We set up random QP of the form \eqref{eq:qp} that have 100 variables
and 100 inequality constraints
in Gurobi and in the serialized and batched versions
of our solver \verb!qpth! and vary the batch size.%
\footnote{Experimental details: we sample
entries of a matrix $U$ from a random uniform distribution and set $Q = U^TU +
10^{-3}I$, sample $G$ with random normal entries, and set $h$ by
selecting generating some $z_0$ random normal and $s_0$ random uniform and
setting $h = Gz_0 + s_0$ (we didn't include equality constraints just for
simplicity, and since the number of inequality constraints in the primary
driver of complexity for the iterations in a primal-dual interior point
method). The choice of $h$ guarantees the problem is feasible.}

\begin{figure}[t]
  \centering
  \includegraphics[width=0.45\textwidth]{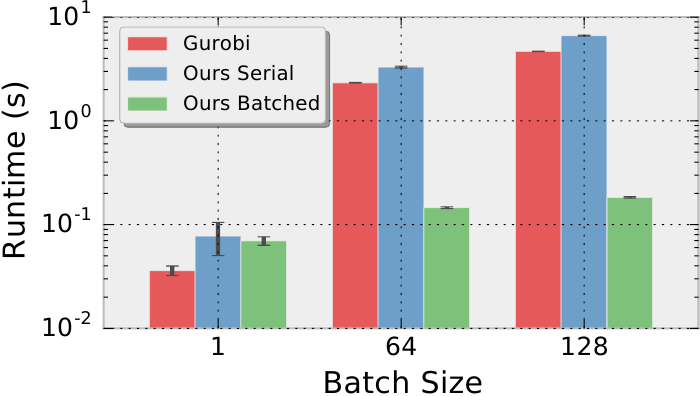}
  \caption{Performance of Gurobi and our QP solver.}
  \label{fig:qp-perf-gurobi}
\end{figure}

Figure~\ref{fig:qp-perf-gurobi} shows the means and standard deviations
of running each trial 10 times, showing that our batched solver
outperforms Gurobi, itself a highly tuned solver for reasonable batch sizes.
For the minibatch size of 128, we solve all problems in an average of 0.18
seconds, whereas Gurobi tasks an average of 4.7 seconds.  In the context of
training a deep architecture this type of speed difference for a single
minibatch can make the difference between a practical and a completely unusable
solution.

\subsection{Total variation denoising}
Our next experiment studies how we can use the OptNet architecture to
\emph{improve} upon signal processing techniques that currently use convex
optimization as a basis.  Specifically, our goal in this case is to denoise a noisy
1D signal given training data consistency of noisy and clean signals generated
from the same distribution.  Such problems are often addressed by convex
optimization procedures, and (1D) total variation denoising is a particularly
common and simple approach.  Specifically, the total variation denoising
approach attempts to smooth some noisy observed signal $y$ by solving the
optimization problem
\begin{equation}
  \argmin_z \frac{1}{2} ||y-z|| + \lambda ||Dz||_1
  \label{eq:tv}
\end{equation}
where $D$ is the first-order differencing operation
expressed in matrix form with rows $D_i=e_i-e_{i+1}$.
Penalizing the $\ell_1$
norm of the signal \emph{difference} encourages this difference to be sparse,
i.e., the number of changepoints of the signal is small, and we end
up approximating $y$ by a (roughly) piecewise constant function.

To test this approach and competing ones on a denoising task, we generate piecewise
constant signals (which are the desired outputs of the learning algorithm) and
corrupt them with independent Gaussian noise (which form the inputs to the
learning algorithm). 
Table~\ref{tab:den} shows the error rate of these
four approaches.

\subsubsection{Baseline: Total variation denoising}
To establish a baseline for denoising performance with total variation, we run
the above optimization problem varying values of $\lambda$ between 0 and 100.
The procedure performs best with a choice of $\lambda \approx 13$,
and achieves a minimum test MSE on our task of about
16.5 (the units here are unimportant, the only relevant quantity is the
relative performances of the different algorithms).

\subsubsection{Baseline: Learning with a fully-connected neural network}
An alternative approach to denoising is by learning from data.
A function $f_\theta(x)$ parameterized by $\theta$ can
be used to predict the original signal.
The optimal $\theta$ can be learned by using
the mean squared error between the true and predicted signals.
Denoising is typically a difficult function to learn and
Table~\ref{tab:den} shows that a fully-connected neural network perform
substantially worse on this denoising task than the
convex optimization problem.
Section~\ref{sec:appendix:den} shows the convergence of the
fully-connected network.

\subsubsection{Learning the differencing operator}
\begin{figure}[t]
  \centering
  \includegraphics[width=0.50\textwidth]{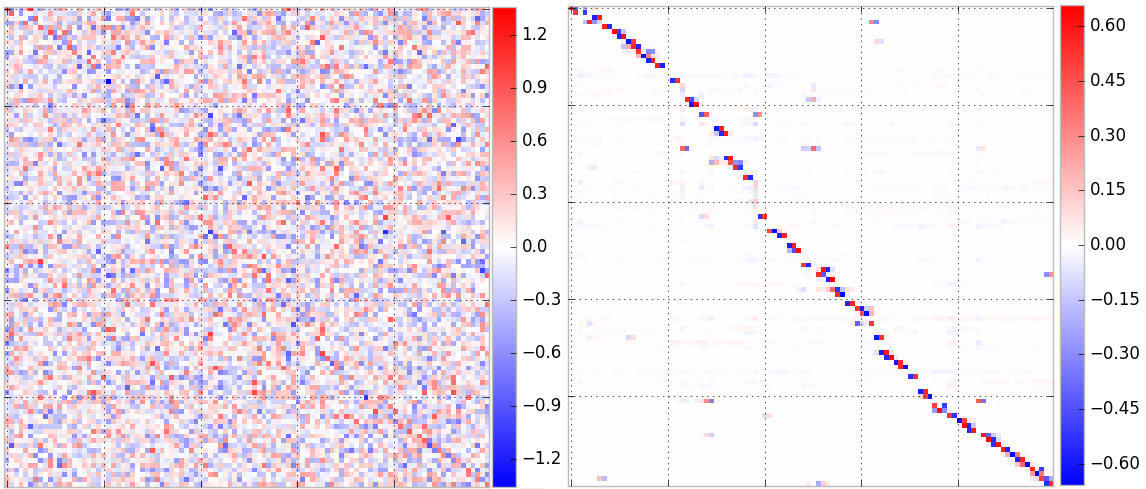}
  \caption{Initial and learned difference operators for denoising.}
  \label{fig:denoiseE:D}
\end{figure}

Between the feedforward neural network approach and the convex total variation
optimization, we could instead use a generic OptNet layers that effectively
allowed us to solve \eqref{eq:tv} using \emph{any} denoising matrix, which we
randomly initialize.  While the accuracy here is substantially lower
than even the fully connected case, this is largely the result of learning an
over-regularized solution to $D$.  This is indeed a point that should be
addressed in future work (we refer back to our comments in the previous section
on the potential challenges of training these layers), but the point we want to
highlight here is that the OptNet layer seems to be learning something very
interpretable and understandable.  Specifically, Figure~\ref{fig:denoiseE:D}
shows the $D$ matrix of our solution before and after learning (we permute the
rows to make them ordered by the magnitude of where the large-absolute-value
entries occurs).  What is interesting in this picture is that the learned $D$
matrix typically captures exactly the same intuition as the $D$ matrix used by
total variation denoising: a mainly sparse matrix with a few entries of
alternating sign next to each other.  This implies that for the data set we
have, total variation denoising is indeed the ``right'' way to think about
denoising the resulting signal, but if some other noise process were to generate
the data, then we can learn that process instead.  We can then attain lower
actual error for the method (in this case similar though slightly higher than
the TV solution), by fixing the learned sparsity of the $D$ matrix and then fine
tuning.

\begin{table}[t]
\begin{center}
\begin{tabular}{lll}
Method & Train MSE & Test MSE \\ \hline
FC Net & 18.5 & 29.8 \\
Pure OptNet & 52.9 & 53.3 \\
Total Variation & 16.3 & 16.5 \\
OptNet Tuned TV & 13.8 & \textbf{14.4}
\end{tabular}
\caption{Denoising task error rates.}
\label{tab:den}
\end{center}
\end{table}

\subsubsection{Fine-tuning and improving the total variation solution}

To finally highlight the ability of the OptNet methods to \emph{improve} upon
the results of a convex program, specifically tailoring to the data.  Here, we
use the same OptNet architecture as in the previous subsection, but initialize
$D$ to be the differencing matrix as in the total variation solution.
As shown in Table~\ref{tab:den}, the procedure is able
to improve both the training and testing MSE over the TV solution,
specifically improving upon test MSE by 12\%.
Section~\ref{sec:appendix:den} shows the convergence of fine-tuning.

\subsection{MNIST}
\label{sec:mnist}

One compelling use case of an OptNet layer is to learn constraints
and dependencies over the output or latent space of a model.
As a simple example to illustrate that OptNet layers can be
included in existing architectures and that the gradients can
be efficiently propagated through the layer, we show
the performance of a fully-connected feedforward network
with and without an OptNet layer in Section~\ref{sec:appendix:mnist}
in the supplemental material.

\begin{figure}[t]
\begin{center}
\scalebox{1.8}{
\begin{TAB}(e,1mm,1mm){|c|c|c|c|}{|c|c|c|c|}
   &  &  & 3 \\
  1 &  &  &  \\
   &  & 4 &  \\
  4 &  &  & 1 \\
\end{TAB}
\hspace{2mm}
\begin{TAB}(e,1mm,1mm){|c|c|c|c|}{|c|c|c|c|}
  2 & 4 & 1 & 3 \\
  1 & 3 & 2 & 4 \\
  3 & 1 & 4 & 2 \\
  4 & 2 & 3 & 1 \\
\end{TAB}
}
\label{fig-sudoku}
\caption{Example mini-Sudoku initial problem and solution.}
\end{center}
\end{figure}

\subsection{Sudoku}
Finally, we present the main illustrative example of the representational power
of our approach, the task of learning the game of Sudoku.  Sudoku is a popular
logical puzzle, where a (typically 9x9) grid of points must be arranged given
some initial point, so that each row, each column, and each 3x3 grid of points
must contain one of each number 1 through 9.  We consider the simpler
case of 4x4 Sudoku puzzles, with numbers 1 through 4, as shown in
Figure~\ref{fig-sudoku}.

Sudoku is fundamentally a constraint satisfaction problem, and is trivial for
computers to solve when told the rules of the game.  However, if we do not know
the rules of the game, but are only presented with examples of unsolved and the
corresponding solved puzzle, this is a challenging task.  We consider this to be
an interesting benchmark task for algorithms that seek to capture complex
strict relationships between all input and output variables.  The input to the
algorithm consists of a 4x4 grid (really a 4x4x4 tensor with a one-hot
encoding for known entries an all zeros for unknown entries), and the desired
output is a 4x4x4 tensor of the one-hot encoding of the solution.

This is a problem where traditional neural networks have difficulties learning
the necessary hard constraints.
As a baseline inspired by the models at
\url{https://github.com/Kyubyong/sudoku},
we implemented a multilayer feedforward network to attempt to solve
Sudoku problems. Specifically, we report results for a
network that has 10 convolutional layers with 512 3x3 filters each,
and tried other architectures as well.
The OptNet layer we use on this task is a completely generic QP in
``standard form'' with only positivity inequality constraints but an
arbitrary constraint matrix $Ax = b$, a small
$Q=0.1I$ to make sure the problem is strictly feasible, and with the linear term
$q$ simply being the input one-hot encoding of the Sudoku problem.
We know that Sudoku \emph{can} be approximated well with a linear program
(indeed, integer programming is a typical solution method for such problems),
but the model here is told nothing about the rules of Sudoku.

\begin{figure}[t]
  \centering
  \hspace{10mm}\includegraphics[width=0.40\textwidth]{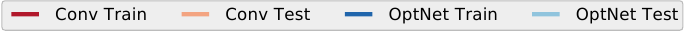}
  \vspace{3mm} \\
  \includegraphics[width=0.45\textwidth]{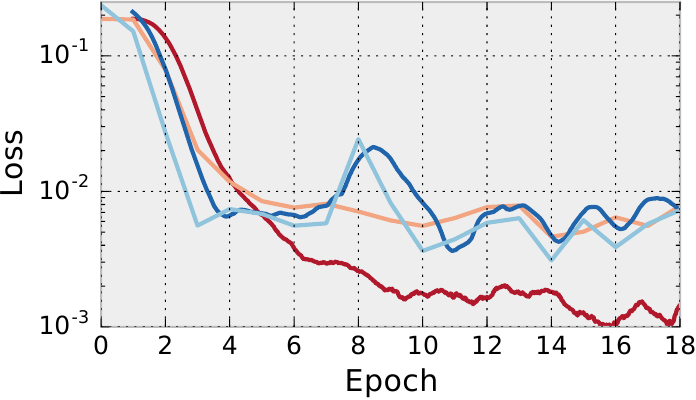}
  \includegraphics[width=0.45\textwidth]{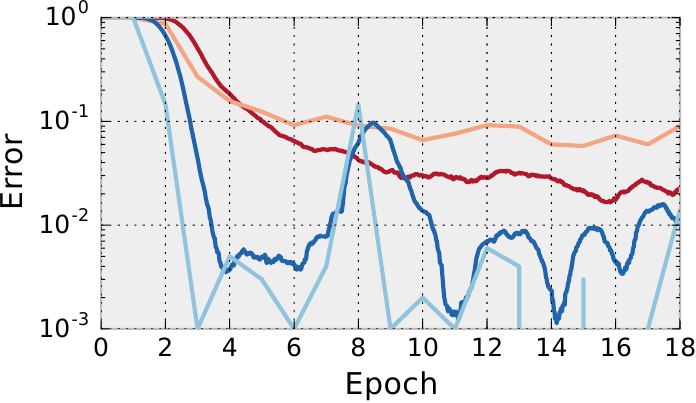}
  \caption{Sudoku training plots.}
  \label{fig:sudoku:convergence}
\end{figure}

We trained these models using ADAM \cite{kingma2014adam} to
minimize the MSE (which we refer to as ``loss'') on a dataset
we created consisting of 9000 training puzzles, and we then tested
the models on 1000 different held-out puzzles.
The error rate is the percentage of puzzles solved correctly
if the cells are assigned to whichever index is largest in the prediction.
Figure~\ref{fig:sudoku:convergence} shows that the convolutional is
able to learn all of the necessary logic for the task and ends up
over-fitting to the training data.
We contrast this with the performance of the OptNet network,
which learns most of the correct hard constraints
and is able to generalize much better to
unseen examples.

\section{Conclusion}
We have presented OptNet, a neural network architecture where we use
optimization problems as a single layer in the network.  We have derived the
algorithmic formulation for differentiating through these layers, allowing for
backpropagating in end-to-end architectures.  We have also developed an
efficient batch solver for these optimizations based upon a primal-dual interior
point method, and developed a method for attaining the necessary gradient
information ``for free'' from this approach.  Our experiments highlight the
potential power of these networks, showing that they can solve problems where
existing networks are very poorly suited, such as learning Sudoku problems
purely from data.
These models add another important primitive to the toolbox
of neural network practitioners and enable many
possible future directions of research of differentiating
through optimization problems.

\newpage
\section*{Acknowledgments}
BA is supported by the National Science Foundation Graduate Research Fellowship
Program under Grant No. DGE1252522. We would like to thank the developers of
PyTorch for helping us add core features, particularly Soumith Chintala and Adam Paszke.
We also thank
Ian Goodfellow,
Yeshu Li,
Lekan Ogunmolu,
Rui Silva,
Po-Wei Wang,
Eric Wong,
and
Han Zhao
for invaluable comments,
as well as Rocky Duan who helped us improve our feedforward network
baseline on mini-Sudoku.

\bibliography{optnet}

\begin{thebibliography}{36}
\providecommand{\natexlab}[1]{#1}
\providecommand{\url}[1]{\texttt{#1}}
\expandafter\ifx\csname urlstyle\endcsname\relax
  \providecommand{\doi}[1]{doi: #1}\else
  \providecommand{\doi}{doi: \begingroup \urlstyle{rm}\Url}\fi

\bibitem[Amos et~al.(2017)Amos, Xu, and Kolter]{amos2017input}
Amos, Brandon, Xu, Lei, and Kolter, J~Zico.
\newblock Input convex neural networks.
\newblock In \emph{Proceedings of the International Conference on Machine
  Learning}, 2017.

\bibitem[Belanger \& McCallum(2016)Belanger and
  McCallum]{belanger2016structured}
Belanger, David and McCallum, Andrew.
\newblock Structured prediction energy networks.
\newblock In \emph{Proceedings of the International Conference on Machine
  Learning}, 2016.

\bibitem[Belanger et~al.(2017)Belanger, Yang, and McCallum]{belanger2017end}
Belanger, David, Yang, Bishan, and McCallum, Andrew.
\newblock End-to-end learning for structured prediction energy networks.
\newblock In \emph{Proceedings of the International Conference on Machine
  Learning}, 2017.

\bibitem[Bertsekas(1999)]{bertsekas1999nonlinear}
Bertsekas, Dimitri~P.
\newblock \emph{Nonlinear programming}.
\newblock Athena scientific Belmont, 1999.

\bibitem[Bonnans \& Shapiro(2013)Bonnans and Shapiro]{bonnans2013perturbation}
Bonnans, J~Fr{\'e}d{\'e}ric and Shapiro, Alexander.
\newblock \emph{Perturbation analysis of optimization problems}.
\newblock Springer Science \& Business Media, 2013.

\bibitem[Boyd \& Vandenberghe(2004)Boyd and Vandenberghe]{boyd2004convex}
Boyd, Stephen and Vandenberghe, Lieven.
\newblock \emph{Convex optimization}.
\newblock Cambridge university press, 2004.

\bibitem[Brakel et~al.(2013)Brakel, Stroobandt, and
  Schrauwen]{brakel2013training}
Brakel, Phil{\'e}mon, Stroobandt, Dirk, and Schrauwen, Benjamin.
\newblock Training energy-based models for time-series imputation.
\newblock \emph{Journal of Machine Learning Research}, 14\penalty0
  (1):\penalty0 2771--2797, 2013.

\bibitem[Chen et~al.(2015)Chen, Schwing, Yuille, and Urtasun]{chen2015learning}
Chen, Liang-Chieh, Schwing, Alexander~G, Yuille, Alan~L, and Urtasun, Raquel.
\newblock Learning deep structured models.
\newblock In \emph{Proceedings of the International Conference on Machine
  Learning}, 2015.

\bibitem[Clarke(1975)]{clarke1975generalized}
Clarke, Frank~H.
\newblock Generalized gradients and applications.
\newblock \emph{Transactions of the American Mathematical Society},
  205:\penalty0 247--262, 1975.

\bibitem[Domke(2012)]{domke2012generic}
Domke, Justin.
\newblock Generic methods for optimization-based modeling.
\newblock In \emph{AISTATS}, volume~22, pp.\  318--326, 2012.

\bibitem[Dontchev \& Rockafellar(2009)Dontchev and
  Rockafellar]{dontchev2009implicit}
Dontchev, Asen~L and Rockafellar, R~Tyrrell.
\newblock Implicit functions and solution mappings.
\newblock \emph{Springer Monogr. Math.}, 2009.

\bibitem[Duchi et~al.(2008)Duchi, Shalev-Shwartz, Singer, and
  Chandra]{duchi2008efficient}
Duchi, John, Shalev-Shwartz, Shai, Singer, Yoram, and Chandra, Tushar.
\newblock Efficient projections onto the l 1-ball for learning in high
  dimensions.
\newblock In \emph{Proceedings of the 25th international conference on Machine
  learning}, pp.\  272--279, 2008.

\bibitem[Fiacco \& Ishizuka(1990)Fiacco and Ishizuka]{fiacco1990sensitivity}
Fiacco, Anthony~V and Ishizuka, Yo.
\newblock Sensitivity and stability analysis for nonlinear programming.
\newblock \emph{Annals of Operations Research}, 27\penalty0 (1):\penalty0
  215--235, 1990.

\bibitem[Goodfellow et~al.(2013)Goodfellow, Mirza, Courville, and
  Bengio]{goodfellow2013multi}
Goodfellow, Ian, Mirza, Mehdi, Courville, Aaron, and Bengio, Yoshua.
\newblock Multi-prediction deep boltzmann machines.
\newblock In \emph{Advances in Neural Information Processing Systems}, pp.\
  548--556, 2013.

\bibitem[Goodfellow et~al.(2014)Goodfellow, Pouget-Abadie, Mirza, Xu,
  Warde-Farley, Ozair, Courville, and Bengio]{goodfellow2014generative}
Goodfellow, Ian, Pouget-Abadie, Jean, Mirza, Mehdi, Xu, Bing, Warde-Farley,
  David, Ozair, Sherjil, Courville, Aaron, and Bengio, Yoshua.
\newblock Generative adversarial nets.
\newblock In \emph{Advances in Neural Information Processing Systems}, pp.\
  2672--2680, 2014.

\bibitem[Gould et~al.(2016)Gould, Fernando, Cherian, Anderson, Santa~Cruz, and
  Guo]{gould2016differentiating}
Gould, Stephen, Fernando, Basura, Cherian, Anoop, Anderson, Peter, Santa~Cruz,
  Rodrigo, and Guo, Edison.
\newblock On differentiating parameterized argmin and argmax problems with
  application to bi-level optimization.
\newblock \emph{arXiv preprint arXiv:1607.05447}, 2016.

\bibitem[Griewank \& Walther(2008)Griewank and Walther]{griewank2008evaluating}
Griewank, Andreas and Walther, Andrea.
\newblock \emph{Evaluating derivatives: principles and techniques of
  algorithmic differentiation}.
\newblock SIAM, 2008.

\bibitem[Johnson et~al.(2016)Johnson, Duvenaud, Wiltschko, Adams, and
  Datta]{johnson2016composing}
Johnson, Matthew, Duvenaud, David~K, Wiltschko, Alex, Adams, Ryan~P, and Datta,
  Sandeep~R.
\newblock Composing graphical models with neural networks for structured
  representations and fast inference.
\newblock In \emph{Advances in Neural Information Processing Systems}, pp.\
  2946--2954, 2016.

\bibitem[Kennedy \& Chua(1988)Kennedy and Chua]{kennedy1988neural}
Kennedy, Michael~Peter and Chua, Leon~O.
\newblock Neural networks for nonlinear programming.
\newblock \emph{IEEE Transactions on Circuits and Systems}, 35\penalty0
  (5):\penalty0 554--562, 1988.

\bibitem[Kingma \& Ba(2014)Kingma and Ba]{kingma2014adam}
Kingma, Diederik and Ba, Jimmy.
\newblock Adam: A method for stochastic optimization.
\newblock \emph{arXiv preprint arXiv:1412.6980}, 2014.

\bibitem[Kunisch \& Pock(2013)Kunisch and Pock]{kunisch2013bilevel}
Kunisch, Karl and Pock, Thomas.
\newblock A bilevel optimization approach for parameter learning in variational
  models.
\newblock \emph{SIAM Journal on Imaging Sciences}, 6\penalty0 (2):\penalty0
  938--983, 2013.

\bibitem[LeCun et~al.(2006)LeCun, Chopra, Hadsell, Ranzato, and
  Huang]{lecun2006tutorial}
LeCun, Yann, Chopra, Sumit, Hadsell, Raia, Ranzato, M, and Huang, F.
\newblock A tutorial on energy-based learning.
\newblock \emph{Predicting structured data}, 1:\penalty0 0, 2006.

\bibitem[Lillo et~al.(1993)Lillo, Loh, Hui, and Zak]{lillo1993solving}
Lillo, Walter~E, Loh, Mei~Heng, Hui, Stefen, and Zak, Stanislaw~H.
\newblock On solving constrained optimization problems with neural networks: A
  penalty method approach.
\newblock \emph{IEEE Transactions on neural networks}, 4\penalty0 (6):\penalty0
  931--940, 1993.

\bibitem[L{\"o}tstedt(1984)]{lotstedt1984numerical}
L{\"o}tstedt, Per.
\newblock Numerical simulation of time-dependent contact and friction problems
  in rigid body mechanics.
\newblock \emph{SIAM journal on scientific and statistical computing},
  5\penalty0 (2):\penalty0 370--393, 1984.

\bibitem[Magnus \& Neudecker(1988)Magnus and Neudecker]{magnus1988matrix}
Magnus, X and Neudecker, Heinz.
\newblock Matrix differential calculus.
\newblock \emph{New York}, 1988.

\bibitem[Mairal et~al.(2012)Mairal, Bach, and Ponce]{mairal2012task}
Mairal, Julien, Bach, Francis, and Ponce, Jean.
\newblock Task-driven dictionary learning.
\newblock \emph{IEEE Transactions on Pattern Analysis and Machine
  Intelligence}, 34\penalty0 (4):\penalty0 791--804, 2012.

\bibitem[Mattingley \& Boyd(2012)Mattingley and Boyd]{mattingley2012cvxgen}
Mattingley, Jacob and Boyd, Stephen.
\newblock Cvxgen: A code generator for embedded convex optimization.
\newblock \emph{Optimization and Engineering}, 13\penalty0 (1):\penalty0 1--27,
  2012.

\bibitem[Metz et~al.(2016)Metz, Poole, Pfau, and
  Sohl-Dickstein]{metz2016unrolled}
Metz, Luke, Poole, Ben, Pfau, David, and Sohl-Dickstein, Jascha.
\newblock Unrolled generative adversarial networks.
\newblock \emph{arXiv preprint arXiv:1611.02163}, 2016.

\bibitem[Morari \& Lee(1999)Morari and Lee]{morari1999model}
Morari, Manfred and Lee, Jay~H.
\newblock Model predictive control: past, present and future.
\newblock \emph{Computers \& Chemical Engineering}, 23\penalty0 (4):\penalty0
  667--682, 1999.

\bibitem[Peng et~al.(2009)Peng, Bo, and Xu]{peng2009conditional}
Peng, Jian, Bo, Liefeng, and Xu, Jinbo.
\newblock Conditional neural fields.
\newblock In \emph{Advances in neural information processing systems}, pp.\
  1419--1427, 2009.

\bibitem[Sastry \& Bodson(2011)Sastry and Bodson]{sastry2011adaptive}
Sastry, Shankar and Bodson, Marc.
\newblock \emph{Adaptive control: stability, convergence and robustness}.
\newblock Courier Corporation, 2011.

\bibitem[Schmidt \& Roth(2014)Schmidt and Roth]{schmidt2014shrinkage}
Schmidt, Uwe and Roth, Stefan.
\newblock Shrinkage fields for effective image restoration.
\newblock In \emph{Proceedings of the IEEE Conference on Computer Vision and
  Pattern Recognition}, pp.\  2774--2781, 2014.

\bibitem[Sra et~al.(2012)Sra, Nowozin, and Wright]{sra2012optimization}
Sra, Suvrit, Nowozin, Sebastian, and Wright, Stephen~J.
\newblock \emph{Optimization for machine learning}.
\newblock Mit Press, 2012.

\bibitem[Stoyanov et~al.(2011)Stoyanov, Ropson, and
  Eisner]{stoyanov2011empirical}
Stoyanov, Veselin, Ropson, Alexander, and Eisner, Jason.
\newblock Empirical risk minimization of graphical model parameters given
  approximate inference, decoding, and model structure.
\newblock In \emph{AISTATS}, pp.\  725--733, 2011.

\bibitem[Tappen et~al.(2007)Tappen, Liu, Adelson, and
  Freeman]{tappen2007learning}
Tappen, Marshall~F, Liu, Ce, Adelson, Edward~H, and Freeman, William~T.
\newblock Learning gaussian conditional random fields for low-level vision.
\newblock In \emph{Computer Vision and Pattern Recognition, 2007. CVPR'07. IEEE
  Conference on}, pp.\  1--8. IEEE, 2007.

\bibitem[Zheng et~al.(2015)Zheng, Jayasumana, Romera-Paredes, Vineet, Su, Du,
  Huang, and Torr]{zheng2015conditional}
Zheng, Shuai, Jayasumana, Sadeep, Romera-Paredes, Bernardino, Vineet, Vibhav,
  Su, Zhizhong, Du, Dalong, Huang, Chang, and Torr, Philip~HS.
\newblock Conditional random fields as recurrent neural networks.
\newblock In \emph{Proceedings of the IEEE International Conference on Computer
  Vision}, pp.\  1529--1537, 2015.

\end{thebibliography}
\bibliographystyle{icml2017}

\clearpage
\appendix
\twocolumn[
  \icmltitle{OptNet: Supplementary Material}
  \begin{icmlauthorlist}
  \icmlauthor{Brandon Amos}{}
  \icmlauthor{J.~Zico Kolter}{}
  \end{icmlauthorlist}
  \vskip 0.3in
]
\icmltitlerunning{OptNet: Supplementary Material}

\section{MNIST Experiment}
\label{sec:appendix:mnist}
In this section we consider the integration of QP OptNet layers into a traditional
fully connected network for the MNIST problem.  The results here show only very
marginal improvement if any over a fully connected layer (MNIST, after all, is
very fairly well-solved by a fully connected network, let alone a convolution
network). But our main point of this comparison is simply to illustrate that we
can include these layers within existing network architectures and efficiently
propagate the gradients through the layer.

Specifically we use a
FC600-FC10-FC10-SoftMax fully connected network and compare it to a
FC600-FC10-Optnet10-SoftMax network, where the numbers after each layer indicate
the layer size.
The OptNet layer in this case includes only inequality constraints and
the previous layer is only used in the linear objective term $p(z_i)=z_i$.
To keep $Q\succ 0$, we use a Cholesky factorization $Q=LL^T+\epsilon I$
and directly learn $L$ (without any information from the previous layer).
We also directly learn $A$ and $G$, and to ensure a feasible solution
always exists, we select some learnable $z_0$ and $s_0$ and set
$b=A z_0$ and $h=Gz_0+s_0$.

\begin{figure}[h]
  \centering
  \includegraphics[width=0.35\textwidth]{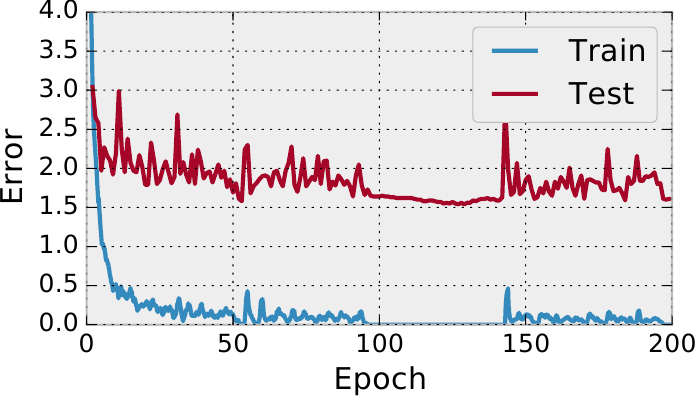}
  \includegraphics[width=0.35\textwidth]{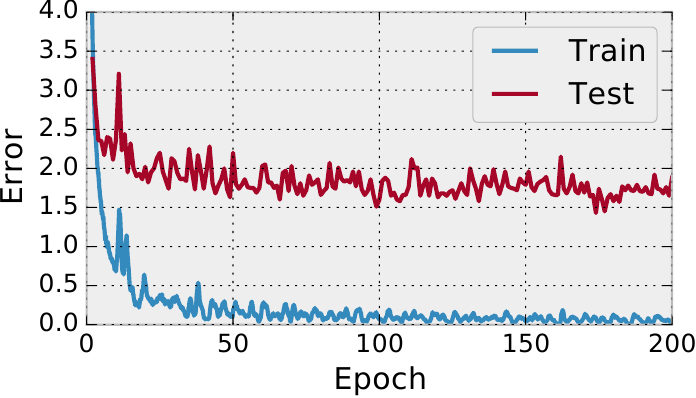}
  \caption{Training performance on MNIST; top: fully connected network;
  bottom: OptNet as final layer.)}
  \label{fig:mnist}
\end{figure}

Figure~\ref{fig:mnist} shows that the results are similar for
both networks with slightly lower error and less variance in
the OptNet network.

\newpage
\section{Denoising Experiment Details}
\label{sec:appendix:den}

\begin{figure}[h]
  \centering
  \includegraphics[width=0.4\textwidth]{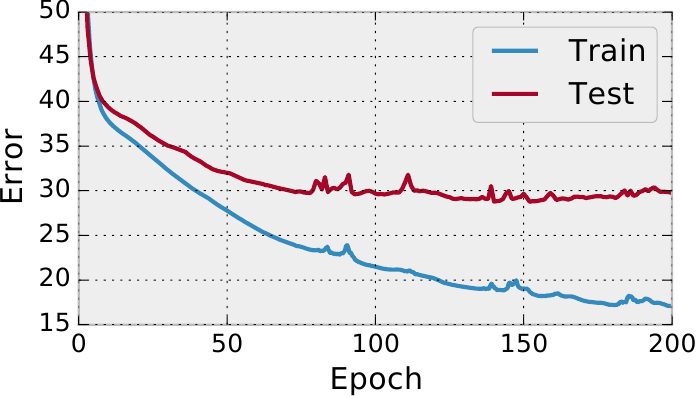}
  \caption{Error of the fully connected network for denoising}
  \label{fig:denoise:fc}
\end{figure}

\begin{figure}[h]
  \centering
  \includegraphics[width=0.35\textwidth]{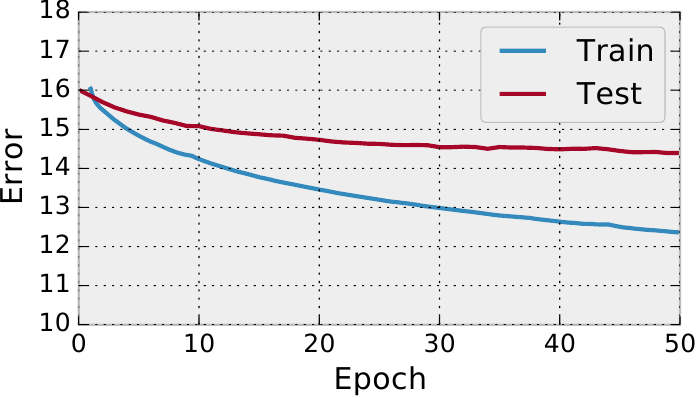}
  \caption{Error rate from fine-tuning the TV solution for denoising}
  \label{fig:denoise:optnet-tv-init}
\end{figure}

Figure~\ref{fig:denoise:fc} shows the error of the fully connected
network on the denoising task and Figure~\ref{fig:denoise:optnet-tv-init} shows the
error of the OptNet fine-tuned TV solution.

\section{Representational power of the QP OptNet layer}

This section contains proofs for those results we highlight in Section
\ref{sec:rep-power}.  As mentioned before, these proofs are all quite
straightforward and follow from well-known properties, but we include them
here for completeness.
\label{sec:appendix:rep-power}

\subsection{Proof of Theorem~\ref{theorem:existence}}
\begin{proof}
  The fact that an OptNet layer is subdifferentiable from strictly convex QPs
  ($Q \succ 0$) follows directly from the well-known result that the solution of a
  strictly convex QP is continuous (though not everywhere differentiable).  Our
  proof essentially just boils down to showing this fact, though we do so by
  explicitly showing that there \emph{is} a unique solution to the Jacobian
  equations \eqref{eq:kkt-diff} that we presented earlier, except on a measure
  zero set.  This measure zero set consists of QPs with degenerate solutions,
  points where inequality constraints can hold with equality yet also have
  zero-valued dual variables.  For simplicity we assume that $A$ has full row
  rank, but this can be relaxed.

  From the complementarity condition, we have that at a primal dual solution
  $(z^\star, \lambda^\star, \nu^\star)$
  \begin{equation}
    \begin{split}
      (Gz^\star - h)_i < 0 & \rightarrow \lambda^\star_i = 0 \\
      \lambda^\star_i > 0 & \rightarrow (Gz^\star - h)_i = 0
    \end{split}
  \end{equation}
  (i.e., we cannot have both these terms non-zero).

  First we consider the (typical) case where exactly one of $(Gz^\star - h)_i$
  and $\lambda^\star_i$ is zero.  Then the KKT differential matrix
  \begin{equation}
    \begin{bmatrix}
      Q & G^T & A^T \\
      D(\lambda^\star)G  & D(Gz^\star-h) & 0 \\
      A & 0 & 0 \\
    \end{bmatrix}
  \end{equation}
  (the left hand side of \eqref{eq:kkt-diff}) is non-singular.  To see this,
  note that if we let $\mathcal{I}$ be the set where $\lambda^\star_i  > 0$,
  then the matrix
  \begin{equation}
    \begin{split}
    & \begin{bmatrix}
      Q & G_{\mathcal{I}}^T & A^T \\
      D(\lambda^\star)G_{\mathcal{I}}  & D(Gz^\star-h)_{\mathcal{I}} & 0 \\
      A & 0 & 0 \\
    \end{bmatrix}
    = \\ & \;\; \begin{bmatrix}
      Q & G_{\mathcal{I}}^T & A^T \\
      D(\lambda^\star)G_{\mathcal{I}}  & 0 & 0 \\
      A & 0 & 0 \\
    \end{bmatrix}
    \end{split}
  \end{equation}  
  is non-singular (scaling the second block by $D(\lambda^\star)^{-1}$ gives a
  standard KKT system \citep[Section 10.4]{boyd2004convex}, which is nonsingular
  for invertible $Q$ and $[G_\mathcal{I}^T$ \; $A^T]$ with full column rank,
  which must hold due to our condition on $A$ and the fact that there must be
  less than $n$ total tight constraints at the solution.  Also note that for any
  $i \not \in \mathcal{I}$, only the 
  $D(Gz^\star -h)_{ii}$ term is non-zero for the entire row in the second block
  of the matrix.  Thus, if we want to solve the system
  \begin{equation}
    \begin{bmatrix}
      Q & G_{\mathcal{I}}^T & A^T \\
      D(\lambda^\star)G_{\mathcal{I}}  & D(Gz^\star-h)_{\mathcal{I}} & 0 \\
      A & 0 & 0 \\
    \end{bmatrix}
    \begin{bmatrix}
      z \\ \lambda \\ \nu
    \end{bmatrix} =
    \begin{bmatrix}
      a \\ b \\ c
    \end{bmatrix} 
  \end{equation}
  we simply first set $\lambda_i$ = $b_i / (Gz^\star - h)_i$ for
  $i \not \in \mathcal{I}$ and then solve the nonsingular system
    \begin{equation}
    \begin{bmatrix}
      Q & G_{\mathcal{I}}^T & A^T \\
      D(\lambda^\star)G_{\mathcal{I}}  & 0 & 0 \\
      A & 0 & 0 \\
    \end{bmatrix}
    \begin{bmatrix}
      z \\ \lambda_\mathcal{I} \\ \nu
    \end{bmatrix} =
    \begin{bmatrix}
      a - G^T_{\bar{\mathcal{I}}} \lambda_{\bar{\mathcal{I}}} \\ b_\mathcal{I} \\ c.
    \end{bmatrix} 
  \end{equation}

  Alternatively, suppose that we have both $\lambda^\star_i = 0$ and $(Gz^\star
  - h)_i = 0$.  Then although the KKT matrix is now singular (any row for which
  $\lambda^\star_i = 0$ and $(Gz^\star -  h)_i = 0$ will be all zero), there
  still exists a solution to the system \eqref{eq:kkt-diff}, because the right
  hand side is always in the range of $D(\lambda^\star)$ and so will also be
  zero for these rows.  In this case there will no longer be a \emph{unique}
  solution, corresponding to the subdifferentiable but not differentiable case.
\end{proof}

\subsection{Proof of Theorem~\ref{theorem:pwlinear}}

\begin{proof}
The proof that an OptNet layer can represent any piecewise linear univariate
function relies on the fact that we can represent any such function in
``sum-of-max'' form
\begin{equation}
  f(x) = \sum_{i=1}^k w_i \max\{a_i x + b, 0\}
\end{equation}
where $w_i \in \{-1,1\}$, $a_i,b_i \in \mathbb{R}$ (to do so, simply proceed
left to right along the breakpoints of the function adding a properly scaled
linear term to fit the next piecewise section).  The OptNet layer simply
represents this function directly.

That is, we encode the optimization problem
\begin{equation}
  \begin{split}
  \minimize_{z \in \mathbb{R},t\in \mathbb{R}^k} \;\; &  \|t\|_2^2 + (z - w^T t)^2 \\
  \subjectto \;\; & a_i x + b_i \leq t_i, \;\; i=1,\ldots,k
\end{split}
\end{equation}
Clearly, the objective here is minimized when $z = w^T t$, and $t$ is as small
as possible, meaning each $t$ must either be at its bound $a_i x + b \leq
t_i$ or, if $a_i x + b < 0$, then $t_i = 0$ will be the optimal solution due to
the objective function.  To obtain a multivariate but elementwise function, we
simply apply 
this function to each coordinate of the input $x$.

To see the specific case of a ReLU network, note that the layer
\begin{equation}
  z = \max\{Wx + b, 0\}
\end{equation}
is simply equivalent to the OptNet problem
\begin{equation}
  \begin{split}
  \minimize_{z} \;\; & \|z - Wx - b\|_2^2 \\
  \subjectto \;\; & z \geq 0.
  \end{split}
\end{equation}
\end{proof}

\subsection{Proof of Theorem~\ref{theorem:rep}}

\begin{proof}
The final theorem simply states that a two-layer ReLU network (more
specifically, a ReLU followed by a linear layer, which is sufficient to achieve
a universal function approximator), can often require exponentially many more
units to approximate a function specified by an OptNet layer. That is, we
consider a single-output ReLU network, much like in the previous section, but
defined for multi-variate inputs.
\begin{equation}
  f(x) = \sum_{i=1}^m w_i \max \{ a_i^T x + b, 0\}
\end{equation}

\begin{figure}[t]
  \centering
  \includegraphics[scale=0.6]{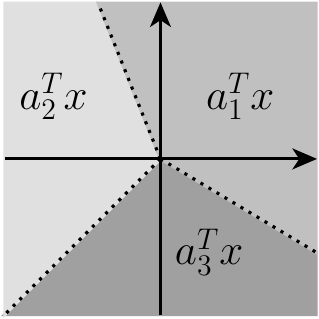}
  \includegraphics[scale=0.6]{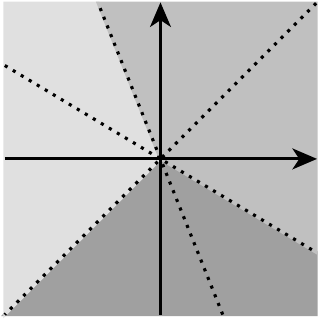}
  \caption{Creases for a three-term pointwise maximum (left), and a ReLU network
    (right).}
  \label{fig:creases}
\end{figure}

Although there are many functions that such a network cannot represent, for
illustration we consider a simple case of a maximum of three linear functions
\begin{equation}
  f'(x) = \max\{a_1^Tx, a_2^T x, a_3^T x\}
\end{equation}
To see why a ReLU is not capable of representing this function exactly, even for
$x \in \mathbb{R}^2$, note that any sum-of-max function, due to the nature of
the term $\max\{a_i^Tx + b_i, 0\}$ as stated above must have ``creases''
(breakpoints in the piecewise linear function), than span the entire input
space; this is in contrast to the max terms, which can have creases that only
partially span the space.  This is illustrated in Figure \ref{fig:creases}.  It
is apparent, therefore, that the two-layer ReLU cannot exactly approximate the
three maximum term (any ReLU network would necessarily have a crease going
through one of the linear region of the original function).  Yet this max
function can be captured by a simple OptNet layer
\begin{equation}
  \begin{split}
    \minimize_z \;\; & z^2  \\
    \subjectto \;\; & a_i^T x \leq z, \; i=1,\ldots, 3.
  \end{split}
\end{equation}

The fact that the ReLU network is a universal function approximator means that
the we \emph{are} able to approximate the three-max term, but to do so means
that we require a dense covering of points over the input space, choose an equal
number of ReLU terms, then choose coefficients such that we approximate the
underlying function on this points; however, for a large enough radius this will
require an exponential size covering to approximate the underlying function
arbitrarily closely.
\end{proof}

Although the example here in this proof is quite simple (and perhaps somewhat
limited, since for example the function can be exactly approximated using a
``Maxout'' network), there are a number of other such functions for which we
have been unable to find any compact representation.  For example, projection of
a point on to the simplex is easily written as the OptNet layer
\begin{equation}
  \begin{split}
  \minimize_{z} \;\; & \|z - x\|_2^2 \\
  \subjectto \;\; & z \geq 0, 1^T z = 1
  \end{split}
\end{equation}
yet it does not seem possible to represent this in closed form as a simple
network: the closed form solution of such a projection operator requires sorting
or finding a particular median term of the data \cite{duchi2008efficient}, which
is not feasible 
with a single layer for any form of network that we are aware of.  Yet for
simplicity we stated the theorem above using just ReLU networks and a
straightforward example that works even in two dimensions.

\end{document}